# DML-RAM: Deep Multimodal Learning Framework for Robotic Arm Manipulation using Pre-trained Models

Sathish Kumar
*Department of Computer Science*
*Cleveland State University*
Cleveland, OH, USA
s.kumar13@csuohio.edu

Swaroop Damodaran
*Department of Computer Science*
*Cleveland State University*
Cleveland, OH, USA
s.damodaran19@vikes.csuohio.edu

Naveen Kumar Kuruba
*Department of Computer Science*
*Cleveland State University*
Cleveland, OH, USA
n.kuruba@vikes.csuohio.edu

Sumit Jha
*Department of Computer Science*
*Florida International University*
Miami, FL, USA
sjha@fiu.edu

Arvind Ramanathan
*Data Science and Learning Division*
*Argonne National Laboratory*
Lemont, IL, USA
ramanathana@anl.gov

*Abstract*—This paper presents a novel deep-learning framework for robotic arm manipulation that leverages multimodal inputs and pre-trained models through a late-fusion learning strategy. Unlike conventional end-to-end or reinforcement learning methods, our core contribution lies in the design of a fusion based framework that integrates separately processed visual and state-based information to improve action prediction. We evaluate our framework on two robotic datasets—BridgeData V2 and Kuka—each comprising diverse manipulation trajectories with sequences of images and corresponding state values, including position, velocity, joint angles, and joint effort. The architecture consists of two independent models: one for extracting features from image sequences using pre-trained networks such as VGG16, ResNet, and Visual Transformers; and another for modeling robot states using algorithms like Random Forests and Gradient Descent. These outputs are fused using our proposed late-fusion mechanism to generate unified r epresentations for predicting continuous action values, which serve as control inputs to drive state transitions. Experimental results confirm the effectiveness of our fusion-based approach across both datasets. The best-performing model—VGG16 combined with a Random Forest Regressor—achieved a mean squared error (MSE) of 0.0021 and a root mean squared error (RMSE) of 0.04604 on BridgeData V2, and an MSE of 0.0028 and RMSE of 0.05123 on the Kuka dataset. Other configurations, s uch a s t he Visual Transformer with Random Forest, performed comparably, highlighting the robustness of our strategy. In addition to its predictive performance, the proposed framework is well-aligned with the goals of human-in-the-loop and adaptive cyber-physical systems, where modularity, interpretability, and multimodal decision-support are essential for real-time collaboration, planning, and control.

*Keywords*—Multimodal fusion, pre-trained models, robotic manipulation, late fusion, cognitive systems, human-in-the-loop, action prediction

## I. INTRODUCTION

Robotic arm manipulation is a fundamental capability required for a wide range of applications, including industrial automation, surgical procedures, assembly lines, and warehouse operations. The ability of a robotic arm to interact with its environment, manipulate objects, and perform precise movements is crucial for enhancing productivity and efficiency across these domains. Traditionally, robotic manipulation has relied on methods grounded in control theory, kinematics, and rule-based strategies. While these approaches have proven effective in structured and predictable environments, they often fall short in complex, dynamic, and unstructured scenarios where the environment or task specifications change frequently.

The emergence of deep learning has introduced a paradigm shift in robotic control, enabling data-driven models to learn from experience and adapt to new situations without explicit programming. Deep learning techniques, particularly those based on neural networks, have shown remarkable success in tasks involving perception, decision-making, and control, making them well-suited for addressing the challenges of robotic manipulation. However, achieving effective manipulation remains a complex problem, especially when dealing with multimodal data, such as visual inputs from cameras and state-based information from sensors embedded in the robotic arm.

In this paper, we propose a novel deep-learning framework for robotic arm manipulation that utilizes pre-trained models for feature extraction, combined with training on a multimodal dataset of image trajectories and corresponding state values to make action predictions. Each image sequence captures the visual observations of the environment, while each state value reflects the arm's physical configuration. Leveraging pre-trained models enables our framework to effectively extract rich features from both visual and state data, following the growing trend toward pre-trained architectures that outperform conventional reinforcement learning methods in scenarios where few-shot or zero-shot learning is preferred. This approach highlights the advantages of multimodal fusion by allowing our model to build on robust, pre-trained representations, which enhances its adaptability and predictive accuracy in complex manipulation tasks. By emphasizing the fusion of visual and state-based information, this work showcases the impact of using pre-trained models in multimodal systems, facilitating more precise and reliable robotic control across a variety of environments and manipulation tasks.

The proposed framework consists of two primary components: the first model extracts features from the time-series data of images, capturing the dynamic visual cues crucial for understanding the environment and the arm's interactions within it. The second model processes the state values, learning to represent the robotic arm's positional and velocity characteristics. By training these models in parallel, we aim to build a comprehensive understanding of both the visual and physical aspects of the manipulation task. The outputs of these models are then fused to create a unified representation that is used to predict the action values (inputs with values for parameters such as joint effort, to a controller which facilitates transitions between states) necessary to control the robotic arm.

The fusion of visual and state-based features in our approach addresses a significant challenge in robotic manipulation — the need for coherent decision-making that integrates both sensory data and mechanical state information. By combining these data modalities, our method enables more accurate predictions of actions, even in scenarios where either the visual data or the state information alone may be insufficient. This multimodal integration is crucial for achieving robust and precise control in diverse operational conditions.

Our framework's contributions include: (1) a novel deep learning architecture that processes multimodal time-series data for robotic control, (2) an efficient fusion mechanism that combines visual features and state-based representations for improved action prediction, and (3) an experimental validation that demonstrates the effectiveness of the proposed approach in enhancing robotic arm manipulation performance. The results highlight the importance of leveraging both image sequences and state values to inform decision-making in real-time, ultimately improving the adaptability and precision of robotic actions.

Beyond traditional robotic applications, the implications of our proposed framework extend to various domains where multimodal sensing is crucial. For example, in autonomous laboratories, robotic systems could use visual and state data fusion to handle delicate procedures and precise manipulations required for experiments. In medical robotics, this approach could enhance the precision of surgical robots by integrating real-time visual feedback and joint state data to ensure accurate and adaptive movements. Additionally, industries like space exploration, agriculture, and manufacturing could benefit from more intelligent, adaptable robots that can process diverse environmental data to make informed decisions.

The remainder of this paper is structured as follows: Section 2 reviews related work on deep learning approaches to robotic manipulation and multimodal data processing. Section 3 details the dataset and the architecture of the proposed framework followed by the experimental setup, including the evaluation metrics used to assess the framework's performance. Section 5 presents a comprehensive analysis of the results, and Section 6 concludes with discussions on the implications of our findings and future research directions.

## II. RELATED WORK

Robotic arm manipulation in dynamic and unstructured environments has been an area of significant research, with advances driven by deep learning and multimodal data fusion. This section reviews notable works that closely relate to the integration of visual and state-based information for robotic control.

Levine et al. [1] demonstrated the effectiveness of using both visual and state data for training robotic policies through deep reinforcement learning. By employing image inputs for visual context and proprioceptive data for the robot's state (such as joint angles and velocities), their method achieved improved performance in complex manipulation tasks. This foundational approach underscores the importance of multimodal fusion, which aligns with our focus on integrating visual and state-based data to inform action predictions more accurately. Gadre et al. [2] presented the CLIP on Wheels (CoW) framework for language-driven zero-shot object navigation. This work underscores the potential of combining visual perception and linguistic cues, showcasing the efficacy of multimodal data fusion. While CoW centers on navigation, the concept of integrating different data types to inform decision-making parallels our strategy of combining visual and state-based features to improve robotic arm manipulation.

Ding et al. [3] introduced the LLM-GROP framework for task and motion planning, integrating large language models (LLMs) for object rearrangement. This approach highlights the utility of multimodal reasoning, incorporating commonsense knowledge with task planning. Although LLM-GROP focuses on language and reasoning, it illustrates the value of combining distinct data sources for planning and control, resonating with our aim to use fused multimodal inputs for precise action prediction.

Ka´roly and Kuti [4] provided a survey categorizing deep learning applications in robotics, emphasizing the role of tailored model architectures in managing complex tasks. Their analysis of modular and end-to-end models highlights the significance of structured model design, supporting our use of specialized architectures for processing visual and state data and fusing outputs to enhance control. This aligns with our novel approach of multimodal data fusion, which improves performance and generalizability in robotic arm manipulation.

TABLE I
SUMMARY OF RELATED WORK IN MULTIMODAL ROBOTIC LEARNING

| Paper Name | Core Idea | Limitations |
|---|---|---|
| Levine et al. [1] | Demonstrated the effectiveness of multimodal data (visual and state inputs) for training robotic policies through deep reinforcement learning, improving manipulation task performance. | Limited to reinforcement learning settings; may not generalize well to unseen scenarios without further pre-training or fine-tuning. |
| Gadre et al. [2] (CLIP on Wheels) | Proposed a framework for language-driven zero-shot object navigation by fusing visual perception and linguistic cues, showcasing multimodal data fusion potential. | Focuses on navigation tasks, limiting applicability to manipulation contexts; reliance on linguistic data. |
| Ding et al. [3] (LLM-GROP) | Integrated large language models with task and motion planning for object rearrangement, combining commonsense knowledge with planning. | Primarily focused on reasoning and language, not manipulation; multimodal inputs are limited to specific tasks. |
| Ka´roly and Kuti [4] | Provided a survey of deep learning in robotics, emphasizing modular and end-to-end architectures for task-specific adaptations. | Survey lacks implementation specifics, limiting direct applicability; does not propose a unified fusion approach. |
| Chen et al. [5] | Introduced a Visuo-Tactile Transformer framework, integrating visual and tactile modalities using transformers to improve manipulation dexterity and robustness. | Focused on tactile-visual fusion; limited consideration of state data and its integration with other modalities. |
| Jetchev and van der Smagt [6] | Developed Visuotactile-RL, leveraging reinforcement learning with high-resolution visual and tactile inputs to improve manipulation precision. | Focused on reinforcement learning, tactile data; less emphasis on multimodal generalization across diverse tasks. |

Chen et al. [5] and Jetchev and van der Smagt [6] explored multimodal approaches to robotic manipulation, combining visual and tactile data.

These works collectively underscore the importance of integrating multimodal data for advanced robotic control. Our framework extends this concept by leveraging visual and state-based data in a unified representation to enhance the fusion process. This results in a more robust and adaptive solution capable of operating effectively in complex, real-world environments.

III. METHODOLOGY

This section outlines the objective of our proposed approach, provides a detailed description of the datasets used, presents the architecture of the multimodal framework, and discusses the experiments conducted to evaluate the model's performance.

The primary objective of this work is to enable robotic arm manipulation using a novel deep learning approach that leverages multimodal data. Our method aims to predict action values for robotic control by integrating information from both visual data (images) and state variables, leading to more precise and adaptable manipulation strategies in complex, real-world environments. The core technical novelty of this work lies in the use of a late-fusion multimodal learning strategy specifically designed to combine visual and state features for action prediction, as opposed to the individual modeling components themselves.

We utilize two datasets in this study: BridgeData V2 and Kuka. The BridgeData V2 dataset [7] is a comprehensive and diverse dataset of robotic manipulation behaviors designed to support research in scalable robot learning. BridgeData V2 is compatible with open-vocabulary, multi-task learning methods, allowing conditioning on goal images or natural language instructions. It is particularly suitable for learning skills that generalize across novel objects and environments. The dataset composition includes 60,096 trajectories in total, from a scripted pick-and-place policy. The data covers 24 environments, grouped into four categories, primarily featuring toy kitchens and diverse tabletop setups, demonstrating 13 distinct skills across varying conditions.

For this study, we focus on 18 out of the 24 available environments. Each trajectory contains around 50 images depicting the task as the robotic arm performs it. Additionally, numerical input variables are included: joint effort, which refers to the forces applied in the six joints of the robotic arm, joint positions, velocities, state values, and desired state values. This diverse dataset supports broad generalization by providing data across various tasks, object configurations, camera poses, and workspace settings, allowing our model to learn manipulation strategies adaptable to different contexts.

The Kuka dataset [14] consists of manipulation trajectories collected from Kuka robotic arms, capturing bin picking and object rearrangement tasks.

Our approach employs a multimodal deep learning framework designed to process both image sequences and numerical state data to predict action values for robotic arm manipulation. The proposed architecture consists of three main components. The first component is the image-based model (Model 1 in Fig. 1), which takes sequences of images from the dataset as input and predicts the future state image based on the past three images. We experiment with different architectures for this task, including pre-trained models such as ResNet, VGG16, and Vision Transformers, to identify the best-performing model for feature extraction from the image data. These pre-trained models were chosen due to their ability to extract effective features from their vast pre-training, though they are not themselves multimodal.

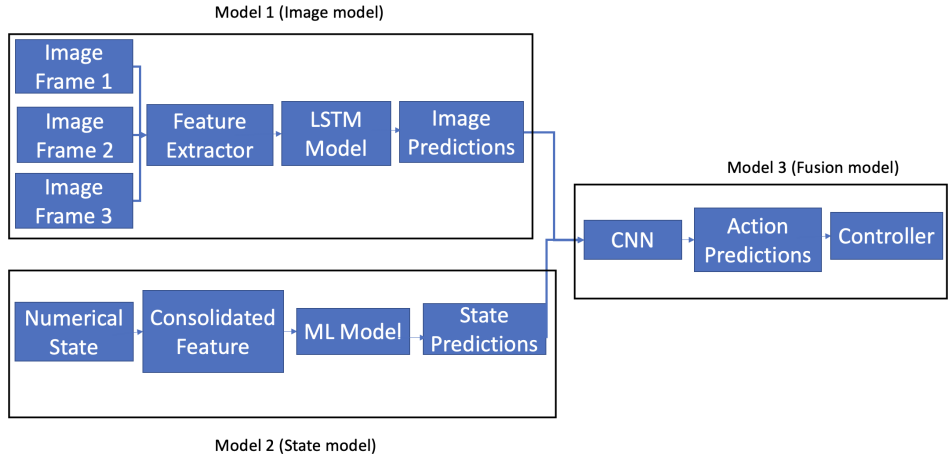

Fig. 1. Architecture of the Proposed Multimodal Framework for Robotic Arm Manipulation.

The second component is the state-based model (Model 2 in Fig. 1), which handles the numerical state variables, predicting future state values using the data related to joint effort, joint positions, joint velocities, and other state variables. We utilize a Random Forest Regressor and a Gradient Descent Algorithm to explore the best fit for this regression task, which were chosen as standard regression models to predict the final action.

The final component is the fusion model (Model 3 in Fig. 1), which combines the outputs from the image-based and state-based models using a late fusion strategy. Unlike early fusion, which combines modalities at the input stage, late fusion processes them independently before merging. The fused representation is then processed by a Convolutional Neural Network (CNN) to predict the action values for the robotic arm. This late-fusion multimodal approach ensures that both visual and state information contribute effectively to predicting robotic actions, improving robustness across different datasets and manipulation contexts.

To evaluate generalization across platforms, we train and validate our models separately on BridgeData V2 and Kuka, comparing performance metrics to understand how well the fusion strategy adapts to different manipulation settings. This comparative analysis provides insights into the model's ability to transfer learned features across different robotic environments.

To evaluate the effectiveness of our proposed multimodal approach, we conducted a series of experiments using different model configurations. The performance of the models is assessed using the following evaluation metrics: Mean Squared Error (MSE), Mean Absolute Error (MAE), and Root Mean Squared Error (RMSE). The experiments include testing the image-based model using ResNet, VGG16, and Vision Transformers to determine the most suitable architecture for feature extraction, as well as evaluating the state-based model's performance using a Random Forest Regressor and a Gradient Descent Algorithm for state value prediction. We also analyze the fusion model's ability to integrate predictions from both the image-based and state-based models to produce accurate action values using a CNN.

All models were trained using a combination of standard optimization techniques and regularization methods to ensure convergence and to avoid overfitting. The experiments were carried out using a training-validation-testing split to rigorously assess model generalization on unseen data.

## IV. RESULTS

The results of our experiments demonstrate the effectiveness of the multimodal fusion approach in predicting action values for robotic manipulation, as illustrated in Figures 2, 3, and 4 (in the Appendix section) and detailed in Tables II, III, IV, V, VI, and VII. The most notable highlight of this study is the performance of the fused model that combined the VGG16 image-based architecture with the Random Forest Regressor for state prediction. This combination achieved the lowest error rates across both datasets, with an MSE of 0.0021, MAE of 0.01311, and RMSE of 0.04604 for BridgeData V2, and an MSE of 0.0028, MAE of 0.01452, and RMSE of 0.05123 for the Kuka dataset. These results highlight the significant impact of multimodal fusion in enabling precise action predictions, which are critical for accurate robotic arm manipulation.

The superior performance of the VGG16 and Random Forest Regressor fusion model can be attributed to the complementary strengths of both components. VGG16, as a convolutional neural network, excels in extracting hierarchical spatial features from visual data, while the Random Forest Regressor effectively handles non-linear dependencies and interactions in state variables. Together, they provide a robust framework for multimodal learning, demonstrating effectiveness across two different datasets with varying task complexities.

The independent performance of the image-based models revealed important differences in their ability to handle visual data. As shown in Table II and Table V, the Visual Transformer achieved the best performance among the image-based models on both datasets, with an MSE of 0.0099 (BridgeData V2) and

TABLE II
PERFORMANCE METRICS FOR MODEL 1: IMAGE BASED MODEL (BRIDGE DATASET)

| Model | MSE | MAE | RMSE |
|---|---|---|---|
| ResNet | 0.0137 | 0.0375 | 0.1171 |
| VGG16 | 0.0134 | 0.0351 | 0.1159 |
| Visual Transformer | 0.0099 | 0.0312 | 0.0995 |

TABLE III
PERFORMANCE METRICS FOR MODEL 2: NUMERICAL STATE VALUES BASED MODEL (BRIDGE DATASET)

| Model | MSE | MAE | RMSE |
|---|---|---|---|
| Random Forest Regressor | 0.00256 | 0.01208 | 0.05064 |
| Gradient Descent Algorithm | 0.00547 | 0.0227 | 0.07399 |

TABLE IV
PERFORMANCE METRICS FOR MODEL 3: FUSION MODEL (BRIDGE DATASET)

| Model Combination | MSE | MAE | RMSE |
|---|---|---|---|
| VGG16 + Random Forest Regressor | 0.0021 | 0.01311 | 0.04604 |
| Visual Transformer + Random Forest Regressor | 0.00215 | 0.0127 | 0.04646 |
| ResNet + Random Forest Regressor | 0.00223 | 0.01290 | 0.04730 |
| VGG16 + Gradient Descent Algorithm | 0.00295 | 0.01785 | 0.05437 |
| Visual Transformer + Gradient Descent | 0.00346 | 0.0179 | 0.05884 |
| ResNet + Gradient Descent Algorithm | 0.00362 | 0.02037 | 0.06022 |

TABLE V
PERFORMANCE METRICS FOR MODEL 1: IMAGE-BASED MODEL (KUKA DATASET)

| Model | MSE | MAE | RMSE |
|---|---|---|---|
| ResNet | 0.0145 | 0.0382 | 0.1204 |
| VGG16 | 0.0141 | 0.0360 | 0.1187 |
| Visual Transformer | 0.0104 | 0.0325 | 0.1021 |

TABLE VI
PERFORMANCE METRICS FOR MODEL 2: NUMERICAL STATE VALUES-BASED MODEL (KUKA DATASET)

| Model | MSE | MAE | RMSE |
|---|---|---|---|
| Random Forest Regressor | 0.00298 | 0.01320 | 0.05287 |
| Gradient Descent Algorithm | 0.00602 | 0.0241 | 0.07834 |

TABLE VII
PERFORMANCE METRICS FOR MODEL 3: NUMERICAL STATE VALUES-BASED MODEL (KUKA DATASET)

| Model Combination | MSE | MAE | RMSE |
|---|---|---|---|
| VGG16 + Random Forest Regressor | 0.0028 | 0.01452 | 0.05123 |
| Visual Transformer + Random Forest Regressor | 0.0029 | 0.01398 | 0.05218 |
| ResNet + Random Forest Regressor | 0.0031 | 0.01475 | 0.05300 |
| VGG16 + Gradient Descent Algorithm | 0.0037 | 0.01821 | 0.05784 |
| Visual Transformer + Gradient Descent | 0.0041 | 0.01908 | 0.06031 |
| ResNet + Gradient Descent Algorithm | 0.0043 | 0.02145 | 0.06250 |

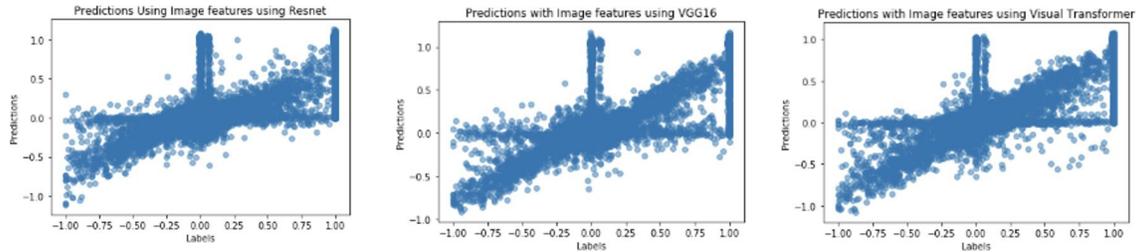

Fig. 2. Image-based model performance with different algorithms.

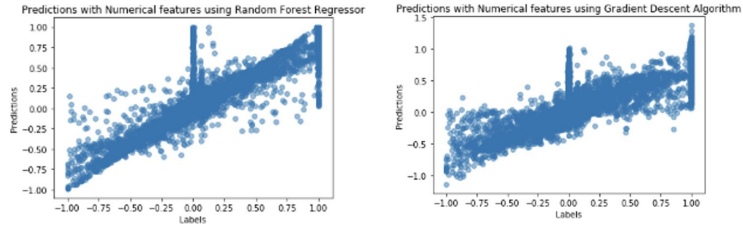

Fig. 3. State-based model performance with different algorithms.

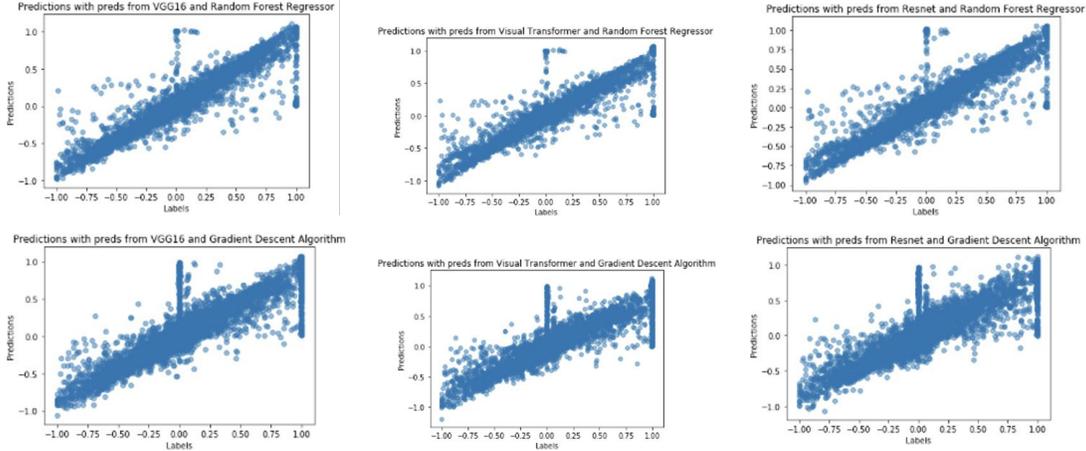

Fig. 4. Performance plots for fusion model with different combinations of algorithms.

0.0104 (Kuka). This indicates that the attention mechanisms in the Transformer are particularly effective in capturing intricate patterns and features in sequential image data. VGG16 followed closely, with slightly higher errors but still competitive results, while ResNet exhibited the highest error rates among the three. The similar trends across the two datasets suggest that the image-based models generalize well across different robotic environments.

For the state-based models, the Random Forest Regressor consistently outperformed the Gradient Descent Algorithm, as shown in Table III and Table VI. The Random Forest Regressor achieved an MSE of 0.00256 (BridgeData V2) and 0.00298 (Kuka), significantly lower than the Gradient Descent Algorithm's MSE of 0.00547 (BridgeData V2) and 0.00602 (Kuka). These results underscore the advantages of using tree-based models for state prediction tasks, particularly in scenarios with non-linear relationships among variables. The slightly higher error rates on the Kuka dataset suggest that state estimation may be more challenging due to variations in force-based interactions.

The fusion models provided deeper insights into the interplay between visual and state-based data. As shown in Tables IV and VII, the VGG16 + Random Forest Regressor combination emerged as the most effective, achieving the lowest error rates across all metrics on both datasets. The Visual Transformer + Random Forest combination also performed well, with an MSE of 0.00215 (BridgeData V2) and 0.0029 (Kuka), slightly trailing behind the VGG16 combination. This consistency in performance highlights the importance of using robust state predictors in multimodal fusion. Interestingly, models that utilized Gradient Descent for state predictions consistently exhibited higher errors compared to those using the Random Forest Regressor. For instance, the VGG16 + Gradient Descent combination recorded an MSE of 0.00295 on BridgeData V2 and 0.0037 on Kuka, significantly higher than the corresponding Random Forest model. This pattern indicates that tree-based methods are better equipped to handle the complexities of state data in multimodal settings.

The performance of ResNet in the fusion models further underscores the importance of selecting appropriate image-based architectures. As shown in Tables IV and VII, ResNet + Random Forest recorded an MSE of 0.00223 (BridgeData V2) and 0.0031 (Kuka), higher than both the VGG16 and Visual Transformer combinations. This finding suggests that ResNet's skip-connection design may not align well with the specific requirements of robotic manipulation tasks, where localized feature extraction and hierarchical representation are critical.

The results emphasize the value of multimodal fusion in enhancing predictive accuracy for robotic manipulation tasks across different datasets. By leveraging pre-trained models for feature extraction, our framework demonstrates the practicality of integrating sophisticated architectures without requiring extensive retraining. This approach aligns with the growing trend of using pre-trained models for few-shot and zero-shot learning, where computational efficiency and adaptability are key considerations. The similar performance trends observed in both datasets indicate that our approach generalizes well to different robotic environments, making it a promising candidate for real-world applications.

In summary, the combination of VGG16 for visual feature extraction with the Random Forest Regressor for state prediction emerged as the most effective fusion model, achieving the best trade-off between computational efficiency and predictive accuracy. The Visual Transformer showed promise as an alternative, particularly for datasets with greater sequential complexity. Future research could explore novel fusion strategies, including hybrid approaches and advanced state predictors, to further optimize the multimodal learning process.

## V. CONCLUSION

In this study, we demonstrated the effectiveness of a multimodal fusion approach that integrates visual and state information for enhancing action prediction in robotic manipulation tasks. The combination of VGG16 with the Random Forest Regressor emerged as the best-performing model across both datasets, achieving an MSE of 0.0021, MAE of 0.01311, and RMSE of 0.04604 on BridgeData V2, and an MSE of 0.0028, MAE of 0.01452, and RMSE of 0.05123 on the Kuka dataset. Although the performance of this model was closely followed by the Visual Transformer + Random Forest Regressor and ResNet + Random Forest Regressor, the results consistently underscore the value of integrating both visual features and state variables to improve the precision of action predictions, which is vital for accurate robotic manipulation.

Our findings suggest that using multimodal data enhances robustness and adaptability in robotic control tasks. By effectively leveraging both visual and numerical state information, our approach provides a comprehensive understanding of the robotic environment, which is essential for predicting actions in dynamic and uncertain conditions. The consistency in performance across two distinct datasets demonstrates the generalizability of the proposed multimodal fusion strategy. The core technical contribution of this work lies in the design of a late-fusion multimodal learning framework specifically tailored for action prediction, enabling the flexible combination of independently processed modalities using lightweight pre-trained models.

The implications of our proposed framework extend to various domains, including autonomous laboratories, medical robotics, space exploration, agriculture, and manufacturing, where precise and adaptive robotic manipulation is essential. Moreover, the model's ability to integrate diverse sensor inputs makes it highly suitable for deployment in human-in-the-loop systems and adaptive cyber-physical environments, where intelligent robots must collaborate with humans and adapt to real-time feedback for decision support.

## VI. FUTURE WORK

Future work will focus on extending this multimodal fusion model by exploring diverse fusion strategies and incorporating additional modalities, such as language, to enhance predictive capabilities in robotic manipulation tasks. Fusion techniques, including early fusion, late fusion, and hybrid fusion, each offer unique advantages depending on the application and data context. Additionally, incorporating explainable AI (XAI) techniques could improve the transparency and trustworthiness of action predictions, especially in safety-critical domains. Implementing interpretable models or post-hoc explainability methods would allow users to understand decision rationales, fostering greater confidence in robotic systems.

Another key direction involves leveraging language as a modality to enhance task understanding and execution. Integrating large language models (LLMs) and multi-task learning frameworks could enable robots to generalize across tasks, improve decision-making, and enhance human-robot interaction, paving the way for more versatile and intelligent systems capable of handling diverse and complex scenarios.